\pdfoutput=1
\documentclass[11pt]{article}

\usepackage[preprint]{acl}

\usepackage{times}
\usepackage{latexsym}

\usepackage[T1]{fontenc}
\usepackage[utf8]{inputenc}

\usepackage{microtype}

\usepackage{inconsolata}

\usepackage{graphicx}
\usepackage{tikz}
\usetikzlibrary{calc, fit}
\usetikzlibrary{plotmarks}
\usepackage{pgfplots}
\pgfplotsset{compat=newest}

\usepackage{amsmath}
\usepackage{amsfonts}
\usepackage{enumitem}
\title{Controlled Low-Rank Adaptation with Subspace Regularization for Continued Training on Large Language Models}

\author{
  \textbf{Yuheng Lu\textsuperscript{1}},
  \textbf{Bingshuo Qian\textsuperscript{1}},
  \textbf{Caixia Yuan\textsuperscript{1}},
  \textbf{Huixing Jiang\textsuperscript{2}},
  \textbf{Xiaojie Wang\textsuperscript{1,}}\thanks{Corresponding author\makeatother}
  \\
  \textsuperscript{1}Beijing University of Posts and Telecommunications,
  \textsuperscript{2}LI Auto Inc.
  \\
  \{yuheng.lu, bsqian, yuancx, xjwang\}@bupt.edu.cn,
  jianghuixing@lixiang.com
}
\begin{document}
\maketitle
\begin{abstract}
  Large language models (LLMs) exhibit remarkable capabilities in natural language processing but
  face catastrophic forgetting when learning new tasks,
  where adaptation to a new domain leads to a substantial decline in performance on previous tasks.
  In this paper,
  we propose Controlled LoRA (CLoRA),
  a subspace regularization method on LoRA structure.
  Aiming to reduce the scale of output change while introduce minimal constraint on model capacity,
  CLoRA imposes constraint on the direction of updating matrix's null space.
  Experimental results on one-stage LLM finetuning tasks and continual learning settings
  highlight the superority of CLoRA as a effective parameter-efficient finetuning method with catastrophic forgetting mitigating.
  Further investigation for model parameters indicates that CLoRA effectively balances the trade-off between model capacity and degree of forgetting.
\end{abstract}

\section{Introduction}
Large language models (LLMs) demonstrate remarkable capabilities in natural language tasks.
However, when performing continued training on additional datasets,
a key challenge may faced, known as catastrophic forgetting \cite{mccloskey1989catastrophic},
where adaptation to a new domain leads to a substantial decline in performance on previous tasks.

Existing approaches to mitigate catastrophic forgetting can be broadly categorized into data-based, architecture-based, and learning-based methods \cite{wangComprehensiveSurveyContinual2023}.
Data-based methods \cite{de2019episodic} primarily based on rehearsing prior training data,
which raises data privacy concerns.
Additionally, for LLMs,
obtaining the necessary prior training data samples is challenging due to their training on massive datas.
Architecture-based methods \cite{wang-etal-2023-rehearsal, razdaibiedina2023progressive} introduce isolated parameters
for each continued training stage for reducing interference.
In contrast, learning-based methods train in the shared vector space,
controlling learning process by adding regularization terms to the loss or employing specific optimization designs.
Inference for architecure-based methods typically involves a selection process \cite{pmlr-v162-gurbuz22a, pmlr-v162-kang22b},
which is more complex than that for learning-based methods.
As continued trained LLMs are generally regarded as foundation models,
flexibility is essential for their broader applications.
Consequently, due to deployment considerations, learning-based methods are preferred over architecture-based methods for LLMs.

The core idea of learning-based methods is to constrain parameter updates,
which aligns precisely with the Parameter-Efficient FineTuning (PEFT) research paradigm of LLMs.
Although initially proposed for computational efficiency,
PEFTs have demonstrated to learn less and forget less\cite{bidermanLoRALearnsLess2024},
primarily due to their constrained model capacity.
Notably, a well-established insight that related to learning-based methods
in PEFT research is that LLMs are primarily finetuned within a specific low-rank subspace,
this insight has led to the development of the Low-Rank Adaptation method (LoRA)\cite{huLoRALowRankAdaptation2021}.

However, LoRA imposes no restrictions on parameter updates beyond the low-rank constraint,
and matrix perturbation theory suggests that even low-rank updates can significantly influence matrix properties
\cite{1370016864442883842, a6ee9c48-1e5b-385a-93c1-cdf3b873de37}.
For instance, in an extreme case, it is theoretically possible to learn a model that eliminates all top-k principal components (optimal rank-k approximation) through a rank-k update,
thus destroy most of the base model's ability.
Therefore, LoRA would be benifit from more constraints for mitigating catastrophic forgetting.
However, more constraints would reduce model capacity for updating,
which influences the effectiveness of training.
For instance, adding L2 regularization significantly restricts the norm of the updating matrix.
Consequently, effective management of the capacity-forgetting balancing has become a major concern.

\begin{figure}[t]
  \centering
  \resizebox{0.85\columnwidth}{!}{
    \begin{tikzpicture}
      \coordinate (A) at (0,0);
      \coordinate (B) at (20pt,20pt);
      \coordinate (X) at (10pt,10pt);
      \draw[blue, -latex, line width=1.5pt] (A) -- (B) node[midway, left=8pt, black] {$x$};

      \node[anchor=west, draw, rectangle, minimum width=30pt, minimum height=20pt, fill=blue!20] (W) at (40pt, 10pt) {$W$};
      \node[right=20pt] (y) at (W.east) {$y$};
      \node[below=5pt] at (y.south) {$+$};

      \coordinate (A1) at (10pt, -35pt);
      \draw[blue, -latex, line width=1.5pt] (A1) -- ($(A1) + (0, 20pt)$) node[left=10pt, midway, black, scale=1, inner sep=0pt] {$x_{\mathrm{Row}(\Delta W)}$};

      \coordinate (A2) at (0, -40pt);
      \draw[blue, -latex, line width=1.5pt] (A2) -- ($(A2) + (20pt, 0)$) node[left=20pt, black, scale=1, inner sep=0pt] {$x_{\mathrm{Null}(\Delta W)}$};

      \node[anchor=west, draw, rectangle, minimum width=30pt, minimum height=20pt, fill=orange] (dW) at (40pt, -33pt) {$\Delta W$};
      \node at ($(dW.south) + (0 , -8pt)$) (perp) {$\perp$};
      \node[anchor=north, draw, dashed, rectangle, minimum width=30pt, minimum height=25pt, fill=blue!20] (P) at ($(dW.south) + (0 , -15pt)$) {$P$};
      \node[anchor=east, scale=0.8] at (P.west) {$\mathrm{Row}(P)\subseteq \mathrm{Null}(\Delta W)$};

      \draw[-latex, shorten <= 0pt] ($(X) + (10pt, 0)$) -- (W.west);
      \draw[-latex, shorten <= 0pt] (W.east) -- (y.west);
      \draw[-latex, shorten <= 0pt] ($(X) + (0, -5pt)$) -- ($(A1) + (0, 22pt)$);

      \draw[-latex, shorten <= 0pt] ($(A1) + (10pt, 8pt)$) -- ($(A1) + (30pt, 8pt)$);
      \draw[-latex, shorten <= 0pt] ($(A2) + (20pt, 0pt)$) -- ($(A2) + (40pt, 0)$);

      \coordinate (A1o) at ($(A1) + (60pt, 8pt)$);
      \node (dy) at ($(A1o) + (30pt, 0)$) {$\Delta y$};
      \draw[-latex, shorten <= 0pt] (A1o) -- (dy);
    \end{tikzpicture}
  }
  \caption{Illustration of the intuition behind our approach.
    For input $x$, the component in $\mathrm{Null}(\Delta W)$ (null space of the updating matrix $\Delta W$) would be ignored,
    the change of output $\Delta y$ is obtained only from the component in $\mathrm{Row}(\Delta W)$ (row space of $\Delta W$, the orthogonal complement of $\mathrm{Null}(\Delta W)$).
    CLoRA introduces a pre-defined subset of $\mathrm{Null}(\Delta W)$ by imposing orthogonal regularization with pre-defined matrix $P$.
    %encourage more $x$ component fall in $\mathrm{Null}(\Delta W)$ than free learned $\Delta W$,
    %and have relatively less influence on its capacity.
    }
  \label{fig:subspace}
\end{figure}
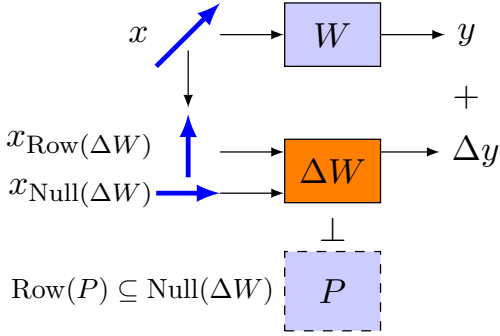

To address this concern, in this work, we propose Controlled LoRA (CLoRA),
a subspace regularization method on LoRA structure.
We start the design of CLoRA from the perspective of the null space of updating matrix.
The intuition behind CLoRA is illustrated in Figure~\ref{fig:subspace},
where the output change $\Delta y$ is derived from applying the updating matrix $\Delta W$
on the component of the input $x$ that falls within the row space of $\Delta W$,
while components in the null space are ignored.
Under this intuition, for reducing the scale of output change,
options include reducing the scale of $\Delta W$,
and encouraging more input component fall in the null space of $\Delta W$.
The former is more related to model capacity,
and for concerns of capacity-forgetting balancing, we focus on the latter.

The dimension of the null space for the updating matrix is directly determined by the rank of it,
which LoRA already addressed.
A key factor remains, the direction of null space,
which influence input components that fall in,
but free-learned LoRA does not constraint.
CLoRA constraint the direction of null space of updating matrix by introducing a pre-defined subspace,
this is implemented by orthogonal regularization with a pre-defined matrix.
Unlike methods that impose restrictions on rank or norm,
which significantly influence model capacity,
CLoRA introduces constraint on the direction of the null space.
We take experiments on commonly used one-stage LLM finetuning evaluations and continual learning evaluations,
results indicate the superiority of CLoRA as an effective approach for parameter-efficient finetuning with catastrophic forgetting mitigating.
Additionally, we take analysis on parameters of the learned model,
results show that CLoRA reduces the scale of output change
with minimal impact on model capacity.

Our contributions are summarized as follows,
\begin{itemize}[leftmargin=*]
  \item We propose CLoRA, a subspace regularization method on LoRA,
        which serves as an advanced parameter-efficient finetuning technique with catastrophic forgetting mitigating for LLMs.
  \item Our proposed CLoRA demonstrates superior performance on both in-domain and out-domain evaluation in commonly used one-stage LLM finetuning setting.
        Additionally, it showns remarkable mitigating of catastrophic forgetting in continual learning setting.
  \item Parameter investigation results indicate that CLoRA effectively balances the trade-off between model capacity and degree of forgetting.
\end{itemize}

\section{Related Works}
\subsection{Mitigating Catastrophic Forgeting}
Catastrophic forgetting is a significant challenge in various transfer learning scenarios,
including continual learning \cite{wangComprehensiveSurveyContinual2023}
and LLM finetuning \cite{wuContinualLearningLarge2024}.
In these settings,
continued training on new tasks may impair abilities of the pre-trained model.
Approaches for mitigating catastrophic forgetting can be broadly categorized into data-based, architecture-based and learning-based methods.

\paragraph{Data-based methods} primarily based on rehearsal of prior training data or representation,
\cite{de2019episodic} introduce an episodic memory for experience rehearsal,
\cite{8100070, chaudhry2019tinyepisodicmemoriescontinual} selects previous training data for rehearsaling.
For LLMs, acquiring the necessary prior training data is challenging due to the extensive amount of data used in their training.
Instead, the concept of rehearsal is commonly adopted by mixing data from general domains for LLM continued training.
This approach is generally orthogonal to model-related methods, thus we will not discuss it further.

\paragraph{Architecture-based methods}
\cite{wang-etal-2023-rehearsal, razdaibiedina2023progressive} introduce isolated parameters
for each continued training stage to reduce interference.
\cite{wang-etal-2023-rehearsal} use isolated parameters for each task,
and enables a selecting mechenism during inference.
Progressive Prompts \cite{razdaibiedina2023progressive}
sequentially concatenates prompts for each task with previously learned prompts.
These architecture-based methods generally require specific techniques for inference and continued training,
resulting in a lack of flexibility, particularly in the context of LLMs.

\paragraph{Learning-based methods}
performs continued training in a shared vector space,
controlling learning process by adding regularization term on loss or applying specific optimization designs.
Notabley, O-LoRA \cite{wangOrthogonalSubspaceLearning2023} introduce regularization with previous continual learned parameters for reducing interference in the multi-stage training setting.
Our proposed CLoRA imposes orthogonal regularization similar to O-LoRA,
but the regularization matrix is not restricted to be the previous learned parameter,
thus CLoRA can be used for one-stage continued training whereas O-LoRA not.

\subsection{LoRA and Subspace Tuning}
Parameter-Efficient FineTuning (PEFT) \cite{han2024parameterefficient} aims to tune models with minimizing computational resources,
which is widely used for large-scale models including LLMs.
Among these methods, LoRA \cite{huLoRALowRankAdaptation2021} and its subsequent variants \cite{wangMiLoRAHarnessingMinor2024, liuDoRAWeightDecomposedLowRank2024} learn a low-rank decomposition for updating parameter matrices,
and could be categorized into learning-based continued training method,
which is the focus of our work.

The core insight of LoRA is to tune model within a low-rank subspace,
and with no additional constraints imposed on this tuning subspace.
Some subsequent works delve deeper into the tuning subspace to mitigate catastrophic forgetting for LLM continued training,
MiLoRA \cite{wangMiLoRAHarnessingMinor2024} and PiSSA \cite{mengPiSSAPrincipalSingular2024} use singular value decomposition (SVD)
components of the original parameters for LoRA initialization,
with MiLoRA uses minor components while PiSSA uses major components;
O-LoRA \cite{wangOrthogonalSubspaceLearning2023} introduce orthogonal regularization for each LoRA subspace.
Our proposed CLoRA also falls within this category,
differing from the selection and utilization of the focused subspace.

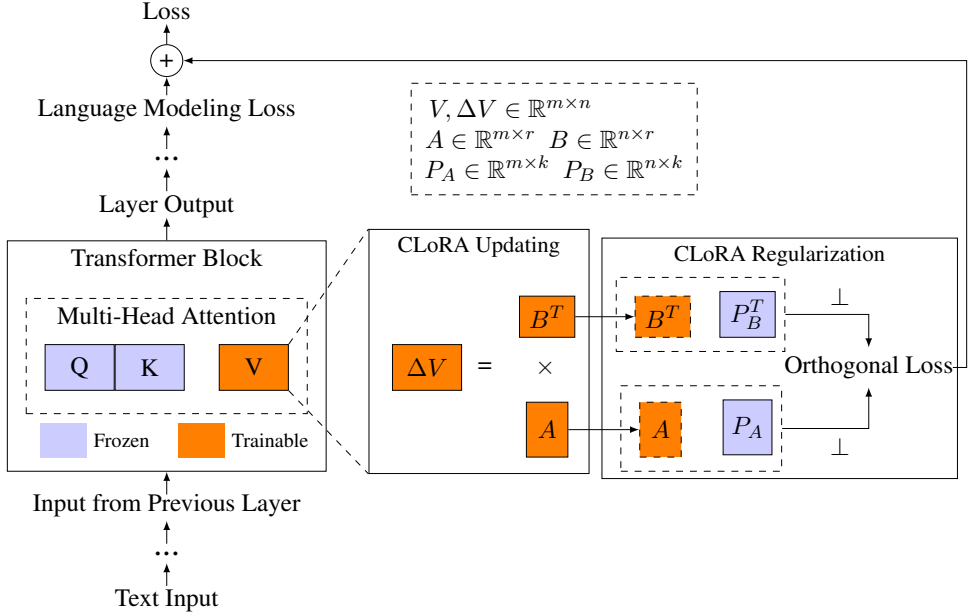
\begin{figure*}[t]
  \centering
  \resizebox{0.75\textwidth}{!}{
    \begin{tikzpicture}
      \tikzset{freeze/.style={fill=blue!20}};
      \tikzset{trainable/.style={fill=orange}};

      \node[draw, rectangle, trainable, minimum width=30pt, minimum height=20pt] (w) at (0,0) {$\Delta V$};
      %\node[anchor=south] at (w.north) {n};
      %\node[anchor=east] at (w.west) {m};

      % transformer
      \node[anchor=east, left=60pt, draw, rectangle, trainable, minimum width=30pt, minimum height=20pt] (V) at (w) {V};
      \node[anchor=east, left=30pt, draw, rectangle, freeze, minimum width=30pt, minimum height=20pt] (K) at (V) {K};
      \node[anchor=east, draw, rectangle, freeze, minimum width=30pt, minimum height=20pt] (Q) at (K.west) {Q};
      \node[draw, dashed, fit=(Q)(K)(V), inner sep=8pt, inner ysep=15pt, yshift=5] (mha) {};
      \node[anchor=north] at (mha.north) {Multi-Head Attention};
      \node[draw, fit=(mha), inner sep=8pt, inner ysep=25pt] (block) {};
      \node[anchor=north] at (block.north) {Transformer Block};
      \node[anchor=south, left=45pt, above=6pt, freeze, minimum width=20pt, minimum height=15pt] (freeze) at (block.south) {};
      \node[anchor=west, scale=0.8] at (freeze.east) {Frozen};
      \node[anchor=south, right=15pt, above=6pt, trainable, minimum width=20pt, minimum height=15pt] (trainable) at (block.south) {};
      \node[anchor=west, scale=0.8] at (trainable.east) {Trainable};

      \node[anchor=north, below=6pt] (inp_p) at (block.south) {Input from Previous Layer};
      \node[anchor=north, below=6pt, scale=1.4] (inp_ell) at (inp_p.south) {...};
      \node[anchor=north, below=6pt] (inp) at (inp_ell.south) {Text Input};
      \draw[-latex, shorten <= -2pt, shorten >= -2pt] (inp) -- (inp_ell);
      \draw[-latex, shorten <= -2pt, shorten >= -2pt] (inp_ell) -- (inp_p);
      \draw[-latex, shorten <= -4pt, shorten >= 0pt] (inp_p) -- (block);

      \node[anchor=south, above=6pt] (out_p) at (block.north) {Layer Output};
      \node[anchor=south, above=6pt, scale=1.4] (out_ell) at (out_p.north) {...};
      \node[anchor=south, above=6pt] (out) at (out_ell.north) {Language Modeling Loss};
      \node[anchor=south, above=6pt, draw, circle, inner sep=2pt] (loss) at (out.north) {+};
      \node[anchor=south, above=7pt] (loss_t) at (loss.north) {Loss};

      \draw[-latex, shorten <= -2pt, shorten >= -2pt] (out_p) -- (out_ell);
      \draw[-latex, shorten <= -2pt, shorten >= -2pt] (out_ell) -- (out);
      \draw[-latex, shorten <= 0pt, shorten >= -4pt] (block.north) -- (out_p);
      \draw[-latex, shorten <= -2pt, shorten >= 0pt] (out) -- (loss);
      \draw[-latex, shorten <= 0pt, shorten >= -2pt] (loss) -- (loss_t);

      \node[anchor=east, above of=w.east, xshift=21pt, align=left, node distance=113pt] (nv) at (w.east) {$V,\Delta V \in \mathbb{R}^{m \times n}$};
      \node[anchor=east, above of=w.east, xshift=34pt, align=left, node distance=100pt] (na) at (w.east) {$A \in \mathbb{R}^{m \times r} ~~B \in \mathbb{R}^{n \times r}$};
      \node[anchor=east, above of=w.east, xshift=40pt, align=left, node distance=87pt] (np) at (w.east) {$P_{A} \in \mathbb{R}^{m \times k} ~~P_{B} \in \mathbb{R}^{n \times k}$};
      \node[draw, dashed, fit=(nv)(na)(np), inner sep=2pt] {};

      \node[anchor=west, right of=w.east, node distance=10pt] (e) at (w.east) {=};

      \node[draw, rectangle, trainable, right=20pt, below=15pt, minimum width=18pt, minimum height=24pt] (A) at (e.east) {$A$};
      %\node[anchor=east] at (A.west) {m};
      %\node[anchor=south] at (A.north) {r};

      \node[right=25pt] (times) at (e.west) {$\times$};

      \node[draw, rectangle, trainable, right=20pt, above=13pt, minimum width=24pt, minimum height=18pt] (B) at (e.east) {$B^{T}$};
      %\node[anchor=east] at (B.west) {r};
      %\node[anchor=south] at (B.north) {n};

      \node[draw, fit=(A)(B)(w), xshift=-2pt, yshift=11pt, inner sep=8pt, inner ysep=18pt] (lora) {};
      \node[anchor=north, scale=0.9] at (lora.north) {CLoRA Updating};

      \node[draw, dashed, rectangle, trainable, right=40pt, minimum width=18pt, minimum height=24pt] (Ap) at (A) {$A$};
      %\node[anchor=east] at (Ap.west) {m};
      %\node[anchor=south] at (Ap.north) {r};

      \draw[dashed] (V.north east) -- (lora.north west);
      \draw[dashed] (V.south east) -- (lora.south west);

      \node[draw, dashed, rectangle, trainable, right=38pt, minimum width=24pt, minimum height=18pt] (Bp) at (B) {$B^{T}$};
      %\node[anchor=east] at (Bp.west) {r};
      %\node[anchor=south] at (Bp.north) {n};

      \node[draw, rectangle, freeze, right=20pt+87pt, below=14pt, minimum width=20pt, minimum height=24pt] (PA) at (e.east) {$P_{A}$};
      %\node[anchor=east] at (PA.west) {m};
      %\node[anchor=south] at (PA.north) {k};

      \node[draw, rectangle, freeze, right=20pt+87pt, above=13pt, minimum width=24pt, minimum height=20pt] (PB) at (e.east) {$P_{B}^{T}$};
      %\node[anchor=east] at (PB.west) {k};
      %\node[anchor=south] at (PB.north) {n};

      \node[right=108pt] (reg) at (times.west) {Orthogonal Loss};

      %\node[draw, dashed, fit=(Ap)(PA), xshift=-5pt, yshift=4pt, inner sep=8pt] (rega) {};
      %\node[draw, dashed, fit=(Bp)(PB), xshift=-5pt, yshift=4pt, inner sep=6pt] (regb) {};
      \node[draw, dashed, fit=(Ap)(PA), xshift=-2pt, inner sep=6pt] (rega) {};
      \node[draw, dashed, fit=(Bp)(PB), xshift=-2pt, inner sep=6pt] (regb) {};

      \node[draw, fit=(rega)(regb)(reg), xshift=-4pt, yshift=7pt, inner sep=2pt, inner ysep=10pt] (clora) {};
      \node[anchor=north, scale=0.9] at (clora.north) {CLoRA Regularization};

      \draw[-latex] (B.east) -- ($(Bp.west)-(0pt,0)$);
      \draw[-latex] (A.east) -- ($(Ap.west)-(0pt,0)$);

      \draw[-latex] (rega.east) -| (reg.south) node[below=8pt, left=8pt, midway, black, scale=1, inner sep=0pt] {$\perp$};
      \draw[-latex] (regb.east) -| (reg.north) node[above=8pt, left=8pt, midway, black, scale=1, inner sep=0pt] {$\perp$};
      \draw[-latex] ($(reg.east)-(4pt,0)$) -- ($(reg.east)+(2pt,0)$) |- (loss.east);

    \end{tikzpicture}
    }
    \caption {Illustration of CLoRA on typical decoder-only transformer based LLMs.
      LoRA updating is applied on v-proj in multi-head attention layer for each layer.
    CLoRA add orthogonal loss computes from trainable LoRA parameters (A and B) to the original language modeling loss. }
    \label{fig:clora}
\end{figure*}

\section{Prelimilaries}
\begin{table}
  \centering
  \begin{tabular}{rl}
    \hline
    Notation & Description \\
    \hline
    $W$ & parameter matrix in base model \\
    $\Delta W$ & updating of the parameter \\
    $x$ & input for $W$ \\
    $y$ & output for $W$, $y=Wx$ \\
    $\Delta y$ & output change, $\Delta y=\Delta Wx$ \\
    $||v||$ & L2 norm of vector $v$ \\
    $||A||$ & L2 norm(largest singular value) \\
    &of matrix $A$ \\
    $r$ & rank of updating matrix \\
    $k$ & number of regularization vectors \\
    \hline
  \end{tabular}
\caption{\label{table:notations}
  Notations.
  }
\end{table}
\subsection{Notations}
The notations commonly used in this paper are summarized in table~\ref{table:notations}.
We provide some additional notes here.
While generally used for denote the input and output of the while model,
we denote $x,y$ as input and output to a single linear layer, represented by $W$.
$||A||$ denotes L2 norm (largest singular value) in our paper, instead of Frobenius norm ($||A||_{F}=\sqrt{\sum A[i,j]^{2}}$).
$r$ and $k$ are most important hyperparameters for CLoRA,
$r$ is the rank of updating matrix, which is used in all LoRA works,
$k$ is the number of regularization vectors(column of regularization matrix) in CLoRA.
\subsection{Problem Definition}
Catastrophic forgetting menifest as performance decline on tasks from previous domain when training on new domain.
In this work, we aim to mitigate catastrophic forgetting in LLM finetuning and continual learning settings.
\subsubsection{LLM Finetuning}
In this setting,
we conduct experiments on one-stage LLM finetuning,
To evaluate this, we conduct both in-domain tasks (demonstrating the effectiveness of training)
and out-domain tasks (from previous domain, indicating the degree of forgetting) for LLM finetuning.
Specifically, we finetune a base LLM on one training dataset,
then take in-domain and out-domain evaluations.
Note that there is no clear domain specific for base LLMs,
but benchmarks exist for evaluating the ability of LLMs on wide range of domains \cite{eval-harness},
and we take those with minimal overlap with training data for out-domain evaluation.

\subsubsection{Continual Learning}
Continual learning focuses on developing learning algorithms to accumulate knowledge on non-stationary data\cite{wangOrthogonalSubspaceLearning2023}.
In this setting,
we conduct experiments for multi-stage finetuning.
Specifically, we finetune the model on a sequence of tasks $D_{1}, \dots, D_{t}$,
where each task $D_{t}$ contains a pair of train and test datasets $D_{t}=(D^{train}_{t}, D^{test}_{t})$.
The $t$-th model with finetuned sequentially on $D^{train}_{1}, \dots, D^{train}_{t-1}$ is tested over all previous test datasets $D^{test}_{1}, \dots, D^{test}_{t-1}$.

\section{Method}
In this section, we introduce Controlled Low-Rank Adaptation (CLoRA) method.
We illustrate the application of CLoRA in transformer-based LLMs in Figure~\ref{fig:clora}.
CLoRA shares the same modeling structure with LoRA,
but imposes on orthogonal regularization term computed using LoRA parameters into the loss function.
\paragraph{CLoRA Modeling}
Consistent with LoRA, CLoRA decomposes the updating for a parameter matrix $W$
to a multiplication of two lor-rank matrices $\Delta W = AB^{T}$,
where $W,\Delta W \in \mathbb{R}^{m\times n}$, $A\in \mathbb{R}^{m\times r}$, $B\in \mathbb{R}^{n\times r}$, $r\ll m,n$.

CLoRA computes orthogonal regularization for $A$ and $B^{T}$ with untrainable pre-defined matrix $P_{A}\in \mathbb{R}^{m\times k}$ and $P_{B}\in \mathbb{R}^{n\times k}$,
where $k$ is a hyperparameter controlling the size of regularization matrix, larger $k$ introduces more constraint.
The orthogonal regularization loss on one LoRA parameter $A$ is defined as
\begin{align}
  L_{orth}(A, P_{A}) = \sum_{i,j}||AP_{A}^{T}[i,j]||^{2}
\end{align}
where $A\in \mathbb{R}^{m\times r}$, $P_{A}\in \mathbb{R}^{m\times k}$.
$L_{orth}(A, P_{A})$ regularize on orthogonality of every $(A[:,i], P_{A}[:,j])$ pairs.
The final loss of CLoRA in a transformer-based LLM is defined as
\begin{align}
  &L_{LM}(\Theta, input) + \\ \notag
  &\lambda\sum_{i}(L_{orth}(A_{i}, P_{A_{i}}) + L_{orth}(B_{i}^{T}, P_{B_{i}}^{T}))
\end{align}
where $L_{LM}(\Theta, input)$ is the original language model loss on text input and LLM parameters $\Theta$,
the summation on $L_{orth}$ is over index of all trainable parameter matrices.
$\lambda$ controls the weighting of orthogonal loss,
we set it to 1 as default.
%We note that practically the orthogonal loss quickly converge to zero, and $\lambda$ is not major.

\paragraph{Initialization}
Following LoRA\cite{huLoRALowRankAdaptation2021}, we initialize $A$ with gaussian noise
and $B$ is zeros, ensuring $\Delta W$ is zero at the begining of training.

For the CLoRA regularization matrices,
following the priciple of Occam's Razor,
we adopt the most simple random initialization here.
For uniform regularization over each row in regularization matrices,
we suggest using orthogonal initialization.
Specifically, for regularization matrix $P\in \mathbb{R}^{m\times k}$, $||P[:,i]||=1$ for every $i$,
and $P[:,i]P[:,j]=0$ for $i\ne j$.
% \begin{itemize}[leftmargin=*]
% \item \textbf{Random Initialize:} Initialize with an orthogonal matrix\cite{saxe2014exactsolutionsnonlineardynamics}.
%   \item \textbf{Initialize from SVD:} Previous works \cite{sharma2024the, wangMiLoRAHarnessingMinor2024} have explored
%         the roles of singular value decomposition (SVD) components in LLM parameters.
%         We adopt this intuition to initialize CLoRA regularization matrices from SVD.
%         For a SVD decomposition of parameter $W=USV^{T}$ with rank $r$, where $W\in \mathbb{R}^{m\times n}$, $U\in\mathbb{R}^{m\times r}$, $S\in\mathbb{R}^{r\times r}$ is a diagonal matrix, $V\in\mathbb{R}^{n\times r}$.
%         For CLoRA updating $\Delta W=AB^{T}$, we initialize the regularization matrix $P_{A}\in \mathbb{R}^{m\times k}$ as $U[:,\mathbf{s}]$, $P_{B}\in \mathbb{R}^{n\times k}$ as $V[:,\mathbf{s}]$,
%         where $\mathbf{s}$ is a list of selecting index with length $k$.
% \end{itemize}

\section{Experiments and Analysis}

\subsection{One-Stage LLM Finetuning}
In this section, we conduct experiments on one-stage LLM finetuning to evaluate our proposed CLoRA
as a parameter-efficient finetuning method.
We aim to answer the following research questions,
\begin{itemize}[leftmargin=*]
  \item \textbf{RQ1:} Does CLoRA performs effectively as a parameter-efficient finetuning method for LLMs with catastrophic forgetting mitigating?
  \item \textbf{RQ2:} How the size of regularization matrix $k$ influence the performance of CLoRA? Does it differs across tasks?
  \item \textbf{RQ3:} How does CLoRA demonstrate superiority on capability-forgetting balancing?
  %\item \textbf{RQ4:} Does CLoRA performs effectively in continual learning setting?
\end{itemize}

% To demonstrate the effectiveness of CLoRA for LLM finetuning (\textbf{RQ1}),
% following previous works on PEFT\cite{liuDoRAWeightDecomposedLowRank2024, wangMiLoRAHarnessingMinor2024},
% we conduct experiments on commonsense reasoning tasks and math tasks.
% We evaluate the model on both in-domain downstream tasks to evaluate finetuning effectiveness
% and on out-domain tasks with minimal overlap with the training dataset to measure the degree of forgetting.
% Different CLoRA regularization matrix size $k$ are tested for each task (\textbf{RQ2}).
% We further take investigation for the parameter of trained model to quantify the capacity-forgetting balancing issue (\textbf{RQ3}).

\subsubsection{Datasets and Tasks}
Following previous works on PEFT\cite{liuDoRAWeightDecomposedLowRank2024, wangMiLoRAHarnessingMinor2024},
we conduct experiments on commonsense reasoning tasks and math tasks.
\paragraph{Commonsense Reasoning Setting}
We use Commonsense170K \cite{hu-etal-2023-llm} for finetuning.
For in-domain evaluation, eight commonsense reasoning datasets are used,
including BoolQ \cite{clark-etal-2019-boolq}, PIQA \cite{Bisk2020}, SIQA \cite{sap-etal-2019-social}, HellaSwag \cite{zellers-etal-2019-hellaswag}, WinoGrande \cite{sakaguchi2019winogrande}, ARC-e, ARC-c \cite{allenai:arc}, and OBQA \cite{mihaylov-etal-2018-suit}.
The tasks are formulated as multiple-choice problem,
and we report accuracy based on the last checkpoint.

For out-domain evaluations, BIG-Bench-Hard \cite{suzgun2022challengingbigbenchtaskschainofthought} and MMLU-Pro \cite{wang2024mmluprorobustchallengingmultitask} are used.
These benchmarks encompass challenging subsets of tasks across a wide range of domains
and are widely employed for evaluating the capabilities of LLMs.
Additionally, they include samples that are more complex than those in our training data, ensuring minimal overlap.
We use lm-eval \cite{eval-harness}, available with MIT License, for reporting out-domain evaluation.

\paragraph{Math Setting}
We use MetaMathQA \cite{yu2024metamath} for finetuning,
which contains 395K samples augmented from the training set of GSM8K \cite{cobbe2021trainingverifierssolvemath} and MATH\cite{hendrycks2021measuring}.
We use test set of GSM8K and MATH for evaluation and report the results on the last checkpoint.

\subsubsection{Comparison Methods}
\begin{itemize}[leftmargin=*]
  \item \textbf{LoRA} \cite{huLoRALowRankAdaptation2021} is a widely-used parameter-efficient finetuning technique, and it serves as the foundation of our proposed CLoRA.
  \item \textbf{DoRA} \cite{liuDoRAWeightDecomposedLowRank2024} is a recent work on structure improvement of LoRA, we include it as a baseline for improved LoRA.
  \item \textbf{PiSSA} \cite{mengPiSSAPrincipalSingular2024} and \textbf{MiLoRA} \cite{wangMiLoRAHarnessingMinor2024} are two variants of LoRA, both employing SVD components for LoRA initialization,
        MiLoRA use minor components while PiSSA use major. Notably, MiLoRA can be categorized as a catastrophic forgetting mitigating method.
  \item \textbf{Reducing the updating rank(-r*):} Lower rank $r$ imposes stricter constraints on the updating matrix.
        We maintain a consistent rank across all methods and consider variations in rank as a separate baseline.
  \item \textbf{L2 regularization(-L2)} introduces L2 regularization for trainable parameters,
        serving as a fundamental approach to limit updates.
  \item \textbf{CLoRA:} Our proposed CLoRA method, with random initialized regularization matrix.
\end{itemize}

\begin{table*}
  \centering
  %\resizebox{0.9\textwidth}{!}{
  \begin{tabular}{lccccccccc|cc}
    \hline
    \multicolumn{10}{c}{In-domain} & \multicolumn{2}{|c}{Out-domain} \\
    Method & BQ & PQ & SQ & HS & WG & ACe & ACc & OQ & Avg. & BBH & MMLU \\
    \hline
    %\multicolumn{10}{c}{\textit{Base Model}} & \multicolumn{2}{|c}{}\\
    %ChatGPT & 73.1 & 85.4 & 68.5 & 78.5 & 66.1 & 89.8 & 79.9 & 74.8 & 77.0 & - & - \\
    LLaMA2-7b & - & - & - & - & - & - & - & - & - &34.91 &  18.56 \\
    \hline
    %\multicolumn{10}{c}{\textit{LoRAs}} & \multicolumn{2}{|c}{}\\
    LoRA & 71.9 & 80.9 & 78.9 & 90.3 & 83.5 & 83.0 & 70.2 & 80.8 & 79.9 & 26.69 &  14.46\\
    DoRA & 73.0 & 81.9 & 80.3 & 90.2 & 82.8 & 84.6 & 69.4 & 81.8 & 80.5 & 28.24 & 11.67\\
    PiSSA & 67.6 & 78.1 & 78.4 & 76.6 & 78.0 & 75.8 & 60.2 & 75.6 & 73.8 & 29.54 & 11.33 \\
    \hline
    %\multicolumn{10}{c}{\textit{Reducing Forgetting}} & \multicolumn{2}{|c}{}\\
    MiLoRA & 71.5 & 82.0 & 80.0 & 91.0 & 83.0 & 82.3 & 68.9 & 81.2 & 80.0 & 25.14 &  17.74\\
    LoRA-r8 & 71.0 & 80.5 & 78.1 & 90.0 & 83.0 & 81.1 & 68.5 & 78.0 & 78.8 & 26.90 &  14.58\\
    LoRA-r16 & 71.0 & 81.8 & 78.9 & 90.3 & 81.1 & 83.1 & 69.7 & 82.2 & 79.8 & 26.73 &  11.54 \\
    LoRA-L2 & 70.3 & 83.0 & 80.2 & 92.7 & 83.1 & 84.2 & 71.2 & 81.4 & 80.8 & 32.93 &  16.59\\
    \hline
    %\multicolumn{10}{c}{\textit{Ours}} & \multicolumn{2}{|c}{}\\
    CLoRA-k128 & 72.7 & 84.1 & 77.7 & 91.6 & 83.0 & 85.3 & 69.9 & 81.6 & 80.7 & 30.82 & 12.07\\
    CLoRA-k256 & 71.3 & 83.2 & 79.1 & 92.4 & 83.2 & 84.5 & 71.0 & 81.0 & 80.7 & 31.92 & 17.81\\
    CLoRA-k512 & 72.8 & 83.0 & 79.5 & 93.0 & 83.9 & 85.7 & 73.0 & 84.8 & 82.0 & 34.32 & 17.00\\
    CLoRA-k1024 & 73.3 & \textbf{84.8} & 79.6 & 91.1 & \textbf{86.1} & 86.9 & 73.1 & 85.6 & 82.6 & 36.49 & 19.52\\
    CLoRA-k2048 & \textbf{73.7} & 84.5 & \textbf{80.9} & \textbf{94.5} & 85.9 & \textbf{88.1} & \textbf{75.9} & \textbf{86.0} & \textbf{83.7} & \textbf{38.67} & \textbf{20.59}\\
    \hline
  \end{tabular}
  %}
\caption{\label{table:in-domain}
  Results for our proposed CLoRA and baselines for in-domain commonsense reasoning evaluations and out-domain LLM benchmarks, with accuracy scores (\%) reported. \textbf{Bold} font indicates the highest performance for each task across all compared PEFT methods.
  }
\end{table*}

\begin{table}[t]
  \centering
  \begin{tabular}{llccc}
    \hline
    Method & GSM8K & MATH\\
    \hline
    LoRA & 60.58 &  16.88\\
    PiSSA & 58.23 &  15.84\\
    MiLoRA & 63.53 &  17.76\\
    \hline
    CLoRA-k64 & 64.29 &  17.52\\
    CLoRA-k128 & \textbf{64.59} &  \textbf{18.38}\\
    CLoRA-k256 & 63.45 &  17.58\\
    \hline
  \end{tabular}

  \caption{\label{table:math}
    Math evaluation on GSM8K and MATH, with accuracy scores (\%) reported.
  }
\end{table}

\subsubsection{Experimental Configuration}
We use the same base LLM choice LLaMA-2-7B \cite{touvron2023llama2openfoundation} and hyperparameter configurations as \cite{wangMiLoRAHarnessingMinor2024}.
Details are listed in Appendix~\ref{appendix:llm}.
Notably, we use 32 (commonsense reasoning) and 64 (math) for updating matrix rank $r$ as default for all methods
if not explicitly specified.
For the size of CLoRA regularization matrix, we select $k$ in $[128, 256, 512, 1024, 2048]$ for commonsense reasoning
and $[64, 128, 256]$ for more challenging math setting.
For LoRA-L2, 1e-5 is used for weighting of L2 regularization,
we note that 1e-4 is also tested, but too large for getting effective finetuning.
We report results finetuned on LLaMA-2-7B here,
more results are listed in Appendix~\ref{appendix:llm}.

\subsubsection{Main Results (RQ1)}
For commonsense reasoning setting,
we report the results of in-domain evaluation and out-domain LLM benchmarks in Table~\ref{table:in-domain}.
The results demonstrate that CLoRA outperforms on all datasets,
surpassing the best baseline for in-domain evaluation by an average accuracy of 2.9 points.
Results for math setting also demonstrate the superority of CLoRA over previous LoRA baselines (Table~\ref{table:math}).

These outcomes suggest that, although primarily proposed for mitigating catastrophic forgetting,
CLoRA also serves as an effective PEFT method.
We attribute this to the nature of LLM finetuning,
which is an instance of transfer learning.
The performance of LLM finetuning is strongly correlated with the base model’s ability,
when catastrophic forgetting occurs during training,
the base model’s strength may diminish.
Therefore, we claim that a method with effective capacity-forgetting balancing
would exhibit strong effectiveness in LLM finetuning.

For out-domain evaluation,
results show that all baselines underperform the base model,
highlighting the severe issue of catastrophic forgetting in this setup.
Notably, our proposed CLoRA not only outperforms all baselines by a significant margin
but also surpasses the base model’s performance.
We attribute this to CLoRA’s effective capacity-forgetting balancing,
which enables the extraction of generally useful knowledge from the commonsense reasoning training dataset.

The superior performance in both in-domain and out-domain evaluations demonstrates that
CLoRA serves as an effective parameter-efficient finetuning method with catastrophic forgetting mitigation.

The superior performance in both in-domain and out-domain evaluations demonstrates that our proposed
CLoRA serves effectively as a parameter-efficient finetuning method with catastrophic forgetting mitigating.
Thus, we answer \textbf{RQ1}.

\subsubsection{Evaluating for different CLoRA $k$ (RQ2)}
The size of the regularization matrix $k$ is a crucial hyperparameter in CLoRA,
balancing the trade-off between model capacity and the degree of forgetting.
We focus here on how $k$ influence the performance of finetuning LLM with CLoRA,
and investigate whether the optimal $k$ is consistent across tasks.

In commonsense reasoning setting,
results show that larger $k$ leads to better performance in both in-domain and out-domain evaluations (Table~\ref{table:in-domain}).
In math setting,
unlike the upward trend in commonsense reasoning setting,
performance decreases when $k$ exceeds 128(Table~\ref{table:math}).
We attribute this discrepancy to the complexity of math tasks,
which require greater model capacity during finetuning.

Emperical results support the intuitive claim
that larger $k$ imposes more restrictions on updates,
which helps mitigating catastrophic forgetting
but potentially limiting finetuning model capacity and harming performance.

Thus, we answer \textbf{RQ2} by demonstrating that the optimal $k$ depends on task complexity.
Notably, our proposed CLoRA provides flexibility in balancing capacity and forgetting by adjusting $k$,
we suggest choosing a smaller $k$ for more challenging tasks.

\subsubsection{Understanding Capacity-Forgetting Balancing(RQ3)}
To answer \textbf{RQ3}, we investigate the parameter of trained models
to quantify the capacity-forgetting balancing issue.

\begin{table}
  \centering
  \begin{tabular}{lccc}
    \hline
    Method & $||\Delta W||$ & $\mathbb{F}$ \\
    \hline
    reference & & 2.42 \\
    \hline
    LoRA & 22.63 & 0.79  \\
    MiLoRA & 24.32 & 0.92  \\
    \hline
    LoRA-r16 & 12.70 & 1.03  \\
    LoRA-r8 & 6.45 & 0.95 \\
    LoRA-L2 & 2.07 & 0.29 \\
    \hline
    CLoRA-k128 & 10.84 & 0.36 \\
    CLoRA-k256 & 10.25 & 0.34  \\
    CLoRA-k512 & 8.19 & 0.27  \\
    CLoRA-k1024 & 6.64 & 0.21 \\
    CLoRA-k2048 & 5.00 & 0.14 \\
    \hline
  \end{tabular}
  \caption{\label{table:rq3}
    Measuring model updating capacity($||\Delta W||$, larger for more capacity) and
    degree of forgetting($\mathbb{F}$, lower for less forgetting) for trained models.
  }
\end{table}

\paragraph{Measuring Model Capacity and Degree of Forgetting}
Consider that catastrophic forgetting primarily arises from output changes caused by parameter updating,
the greater the impact of these updates, the more severe the catastrophic forgetting may be.
We measure the degree of forgetting with the relative scale of output change in the parameter level,
to be specific, for updating matrix $\Delta W$, with input $x$, the relative scale of output change (denoted as $\mathbb{F}$) is defined as
\begin{align}
  \mathbb{F}(\Delta W, x)=\frac{||\Delta Wx||}{||x||}\label{eq:forget}
\end{align}
We highlight the role of $x$ in the measurement of $\mathbb{F}$,
as it reflects the real world case.
Specifically, we sample 100 text data from test set
and collect input $x$ for each parameter from the model forward pass.

To measure model capacity,
we note that there is a gap between theoretical capacity of a model \cite{abu1989vapnik} and
the practical outcome of the learned model.
Therefore, we delegate the measurement of model capacity to the scale of the parameters in the learned model.
Specifically, we measure the L2 norm $||\Delta W||$ for each updating parameter matrix.

\paragraph{Results and Analysis}
We report the measurements averaged over all tokens and all updating parameters in Table~\ref{table:rq3}.
All models use LoRA rank $r$ of 32 unless specified otherwise.
% $\mathbb{F}$ indicates the relative scale of output change (degree of forgetting),
% and $||\Delta W||$ represents our delegated measurement of model capacity.

The ``reference'' row is computed using the LoRA trained model,
noting the output scale of original parameter $W$ instead of $\Delta W$.
Compared with ``reference'' and LoRA, the difference of $\mathbb{F}$ is not far,
suggesting that LoRA training indeed introduces significant output change,
thus still prone to catastrophic forgetting.

For MiLoRA, although intuitively promising, without effective control during training,
it did not mitigate catastrophic forgetting, as evidenced by both downstream evaluations (Table~\ref{table:in-domain})
and the similar $\mathbb{F}$ and $\Delta W$ with LoRA.

For LoRA with lower rank (r8/16),
after training, with $||\Delta W||$ indicates the reduction of capacity, $\mathbb{F}$ does not show a decrease.
Although theoretically, reducing the rank of the update matrix can increase the dimension of the null space and help to reduce the scale of output change,
results not show this case.
This suggests that altering $r$ may not a effctive way to alter forgetting.

For LoRA-L2, $\mathbb{F}$ indicates that it indeed mitigate forgetting,
but in a large cost of capacity, demonstrated by the very small $||\Delta W||$.

For our proposed CLoRA, $\mathbb{F}$ shows a significantly reduce the scale of output change,
while a relatively larger $||\Delta W||$ is maintained.
This indicates that CLoRA minimizes catastrophic forgetting caused by large updates
while having a subtle impact on model capacity.
Thus we answer \textbf{RQ3} that CLoRA performs effectively on capacity-forgetting balancing.

\begin{table*}
  \centering
  %\resizebox{0.85\textwidth}{!}{
  \begin{tabular}{l|cccc|cccc}
    \hline
    & \multicolumn{4}{|c}{Standard CL Benchmark} & \multicolumn{4}{|c}{Large Number of Tasks} \\
    Method & Order-1 & Order-2 & Order-3 & avg. & Order-4 & Order-5 & Order-6 & avg. \\
    \hline
    SeqFT & 18.9 & 24.9 & 41.7 & 28.5 & 7.4 & 7.4 & 7.5 & 7.4\\
    SeqLoRA & 44.6 & 32.7 & 53.7 & 43.7 & 2.3 & 0.6 & 1.9 & 1.6\\
    IncLoRA & 66 & 64.9 & 68.3 & 66.4 & 63.3 & 58.5 & 61.7 & 61.2\\
    \hline
    Replay & 55.2 & 56.9 & 61.3 & 57.8 & 55 & 54.6 & 53.1 & 54.2\\
    \hline
    EWC & 48.7 & 47.7 & 54.5 & 50.3 & 45.3 & 44.5 & 45.6 & 45.1\\
    LwF & 54.4 & 53.1 & 49.6 & 52.3 & 50.1 & 43.1 & 47.4 & 46.9\\
    \hline
    L2P & 60.3 & 61.7 & 61.1 & 60.7 & 57.5 & 53.8 & 56.9 & 56.1\\
    LFPT5 & 67.6 & 72.6 & 77.9 & 72.7 & 70.4 & \textbf{68.2} & 69.1 & 69.2\\
    O-LoRA & 75.4 & 75.7 & 76.3 & 75.8 & \textbf{72.3} & 64.8 & \textbf{71.6} & \textbf{69.6}\\
    \hline
    \textbf{CLoRA} & \textbf{79.7} & \textbf{79.1} & \textbf{78.2} & \textbf{79.0} &70.7 &65.6 &68.2 &68.1\\
    \hline
    PerTaskFT & 70.0 & 70.0 & 70.0 & 70.0 & 78.1 & 78.1 & 78.1 & 78.1 \\
    MTL & 80.0 & 80.0 & 80.0 & 80.0 & 76.5 & 76.5 & 76.5 & 76.5 \\
    \hline
  \end{tabular}
  %}
  \caption{\label{table:olora}
    Results on two CL benchmarks with T5-large base model. Averaged accuracy after training on the last task is reported. \textbf{Bold} font indicates the highest performance across all compared CL methods.
  }
\end{table*}

\subsection{Continual Learning}
\subsubsection{Experimental Setup}
To demonstrate the effectiveness of CLoRA for continual learning(CL) setting,
we conduct experiments on standard CL benchmark and more challenging large number of tasks benchmark,
following the experiment setup of O-LoRA\cite{wangOrthogonalSubspaceLearning2023}.
\paragraph{Datasets and Tasks}
The standard CL benchmark consists of five text classification datasets\cite{zhangCharacterlevelConvolutionalNetworks2015}.
The large number of tasks benchmark consists of 15 datasets \cite{razdaibiedina2023progressive},
include tasks for natural language understanding and text classification.
Task samples follows previous work \cite{wangOrthogonalSubspaceLearning2023}.
Details for tasks are listed in Appendix~\ref{appendix:cont}.
\paragraph{Comparison Methods}
We compare CLoRA with normal finetuning baselines and previous CL methods.
We include non CL results that train separate model for each task (\textbf{PerTaskFT}) and multi-task learning (\textbf{MTL}) as reference.
\begin{itemize}[leftmargin=*]
  \item \textbf{Normal Finetuing} baselines include sequentially training on same parameter space with
        full parameter finetune (\textbf{SeqFT}) and LoRA (\textbf{SeqLoRA}),
        and incremental learning of new LoRA parameters on a sequential series of tasks.
  \item \textbf{Continual Learning} methods include data-based methods
        \textbf{Replay}; architecture-based methods \textbf{L2P}\cite{wangLearningPromptContinual2022}, \textbf{LFPT5}\cite{qin2022lfpt}\textbf{O-LoRA}\cite{wangOrthogonalSubspaceLearning2023};
        and learning-based methods \textbf{EWC}\cite{doi:10.1073/pnas.1611835114}, \textbf{LwF}\cite{leibeComputerVisionECCV2016}.
        Details for these methods are listed in Appendix~\ref{appendix:cont}.
\end{itemize}

\paragraph{Experimental Configuration}
Following O-LoRA\cite{wangOrthogonalSubspaceLearning2023}, we use T5-large as base model,
and finetune on each task with specified order(Appendix~\ref{appendix:cont}).
We train each task with one epoch,
with constant learning rate of 1e-3, batch size of 64, dropout rate of 0.1, weight decay rate of 0,
and LoRA dim $r$ of 8.
CLoRA regularization matrix size $k$ is set to 256.

\subsubsection{Results and Analysis}
We report the results in Table~\ref{table:olora}.
Results demonstrate that CLoRA outperforms all comparision methods,
include the most related strong baseline O-LoRA,
with a notable margin in the stadard CL benchmark.
We attribute this to the advantage of CLoRA toward O-LoRA:
1. CLoRA helps learning in the first finetuning stage while O-LoRA not;
2. CLoRA can independently alter $k$ for balancing learning and forgetting,
while ``$k$'' equivalent in O-LoRA is restrained by LoRA $r$.

In the large number of tasks benchmark, CLoRA performs near the strong baseline O-LoRA and LFPT5.
We note that vanilla CLoRA with random regularization matrix is a learning-based method,
without machanism for isolating finetuning tasks.

\section{Conclusion}
In this paper, we introduce Controlled Low-Rank Adaptation(CLoRA),
a simple yet effective parameter-efficient finetuning method for LLMs
that mitigates catastrophic forgetting.
We investigate the effectiveness of CLoRA on both one-stage LLM finetuning and continual learning settings.
Experiment results demonstrate the effectiveness of CLoRA as a parameter-efficient finetuning method with catastrophic forgetting mitigating.
Further investigation for model parameters indicates that CLoRA effectively balances the trade-off between model capacity and degree of forgetting.

\section{Limitations}
There are still several limitations that we reserve for future work:
1) We use the simplest random initialization for regularization matrix,
insight for more dedicated choice may benifit CLoRA learning,
such as combine CLoRA with architecture-based continual learning method(O-LoRA).
2) We delegate the measurement of model capacity and degree of forgetting to simple measurement of scale.
Although these measurements reveal significant differences between CLoRA and previous works,
we believe that further investigation would aid in the design of methods with stronger capacity-forgetting balancing capability.

\bibliography{custom}

\clearpage
\appendix

\begin{table*}[h]
  \centering
  \begin{tabular}{rcc}
    \hline
    Hyperparameter & ~~~~~~~~~~CS & ~~~~~Math\\
    \hline
    LoRA rank r & ~~~~~~~~~~32 & ~~~~~64\\
    LoRA $\alpha$ & ~~~~~~~~~~64 & ~~~~~128 \\
    Dropout & \multicolumn{2}{c}{0.05}\\
    Optimizer & \multicolumn{2}{c}{AdamW}\\
    LR for LLaMA-2-7B & \multicolumn{2}{c}{3e-4}\\
    LR for LLaMA-3-8B & \multicolumn{2}{c}{1e-4}\\
    LR Scheduler & \multicolumn{2}{c}{Linear}\\
    Batch Size & \multicolumn{2}{c}{16}\\
    Warmup Steps & \multicolumn{2}{c}{100}\\
    Epochs & \multicolumn{2}{c}{3}\\
    LoRA target modules & \multicolumn{2}{c}{query, key, value, MLP up, MLP down}\\
    \hline
  \end{tabular}

  \caption{\label{table:hp}
    Hyperparameters for commonsense reasooning (CS) and Math settings.
  }
\end{table*}

\section{Experiment Details for One-Stage Finetuning}
\label{appendix:llm}
\subsection{Hyperparameter Settings}

Table~\ref{table:hp} shows our detailed hyperparameters. This setting follows MiLoRA\cite{wangMiLoRAHarnessingMinor2024} and DoRA\cite{liuDoRAWeightDecomposedLowRank2024}.

\subsection{Computation Environment}
All of our experiments are conducted on 8 NVIDIA A800 GPUs.
All methods for LoRA subsequents use Huggingface peft library\footnote{\url{https://github.com/huggingface/peft}},
training is conducted using trainer in Huggingface transformers library\footnote{\url{https://github.com/huggingface/transformers}},
with DeepSpeed ZeRO\cite{10.5555/3433701.3433727} intergration.

\subsection{Additional CLoRA variants}
We use the simplest random initialization for CLoRA regularization matrix in the main paper.
Considering the idea of PiSSA and MiLoRA that
explore the roles of singular value decomposition (SVD) components in LLM parameters,
we adopt this intuition to initialize CLoRA regularization matrices from SVD.
For a SVD decomposition of parameter $W=USV^{T}$ with rank $r$,
where $W\in \mathbb{R}^{m\times n}$, $U\in\mathbb{R}^{m\times r}$, $S\in\mathbb{R}^{r\times r}$ is a diagonal matrix, $V\in\mathbb{R}^{n\times r}$.
For CLoRA updating $\Delta W=AB^{T}$, we initialize the regularization matrix $P_{A}\in \mathbb{R}^{m\times k}$ as $U[:,\mathbf{s}]$, $P_{B}\in \mathbb{R}^{n\times k}$ as $V[:,\mathbf{s}]$,
where $\mathbf{s}$ is a list of selecting index with length $k$.
We add two CLoRA variants as follows, and conduct experiments on commonsense reasoning setting,
\begin{itemize}[leftmargin=*]
  \item \textbf{CLoRA-major:} Use SVD major components to initialize CLoRA regularization matrix.
  \item \textbf{CLoRA-minor:} Use SVD minor components to initialize CLoRA regularization matrix.
\end{itemize}

\subsection{Full Results on Commonsense Finetuning}

\begin{table*}
  \begin{center}
    \resizebox{\textwidth}{!}{
  \begin{tabular}{llccccccccc}
    \hline
Model & PEFT & BoolQ & PIQA & SIQA & HS & WG & ARC-e & ARC-c & OBQA & Avg.\\
\hline
ChatGPT & - & 73.1 & 85.4 & 68.5 & 78.5 & 66.1 & 89.8 & 79.9 & 74.8 & 77.0\\
\hline
LLaMA-2-7B & LoRA & 71.9 & 80.9 & 78.9 & 90.3 & 83.5 & 83.0 & 70.2 & 80.8 & 79.9\\
 & PiSSA & 67.6 & 78.1 & 78.4 & 76.6 & 78.0 & 75.8 & 60.2 & 75.6 & 73.8\\
 & MiLoRA & 71.5 & 82.0 & 80.0 & 91.0 & 83.0 & 82.3 & 68.9 & 81.2 & 80.0\\
 & DoRA & 73.0 & 81.9 & 80.3 & 90.2 & 82.8 & 84.6 & 69.4 & 81.8 & 80.5\\
\hline
 & LoRA-r8 & 71.0 & 80.5 & 78.1 & 90.0 & 83.0 & 81.1 & 68.5 & 78.0 & 78.8\\
 & LoRA-r16 & 71.0 & 81.8 & 78.9 & 90.3 & 81.1 & 83.1 & 69.7 & 82.2 & 79.8\\
\hline
 & LoRA-L2-0.0001 & - & - & - & - & - & - & - & - & -\\
 & LoRA-L2-0.00001 & 70.3 & 83.0 & 80.2 & 92.7 & 83.1 & 84.2 & 71.2 & 81.4 & 80.8\\
\hline
 & CLoRA-random-k128 & 72.7 & 84.1 & 77.7 & 91.6 & 83.0 & 85.3 & 69.9 & 81.6 & 80.7\\
 & CLoRA-random-k256 & 71.3 & 83.2 & 79.1 & 92.4 & 83.2 & 84.5 & 71.0 & 81.0 & 80.7\\
 & CLoRA-random-k512 & 72.8 & 83.0 & 79.5 & 93.0 & 83.9 & 85.7 & 73.0 & 84.8 & 82.0\\
 & CLoRA-random-k1024 & 73.3 & 84.8 & 79.6 & 91.1 & \textbf{86.1} & 86.9 & 73.1 & 85.6 & 82.6\\
 & CLoRA-random-k2048 & 73.7 & 84.5 & \textbf{80.9} & 94.5 & 85.9 & \textbf{88.1} & 75.9 & \textbf{86.0} & \textbf{83.7}\\
\hline
 & CLoRA-major-k128 & 72.4 & 81.9 & 77.9 & 83.9 & 82.4 & 84.4 & 70.0 & 82.6 & 79.4\\
 & CLoRA-major-k256 & 73.2 & 83.5 & 79.6 & 93.0 & 83.3 & 88.1 & 72.6 & 84.2 & 82.2\\
 & CLoRA-major-k512 & 73.6 & 83.7 & 79.9 & 93.4 & 83.9 & 86.4 & 73.0 & 86.0 & 82.5\\
 & CLoRA-major-k1024 & 73.2 & \textbf{85.5} & 80.5 & 94.3 & 85.7 & 87.2 & 75.9 & 85.4 & 83.5\\
 & CLoRA-major-k2048 & \textbf{73.9} & 84.8 & 80.6 & \textbf{95.0} & 85.3 & 87.7 & \textbf{76.5} & 84.6 & 83.6\\
\hline
 & CLoRA-minor-k128 & 71.5 & 82.7 & 78.7 & 91.8 & 83.2 & 85.0 & 70.9 & 81.6 & 80.7\\
 & CLoRA-minor-k256 & 72.6 & 83.5 & 80.2 & 91.3 & 85.4 & 85.4 & 72.1 & 83.6 & 81.8\\
 & CLoRA-minor-k512 & 73.0 & 84.0 & 80.1 & 93.1 & 82.0 & 86.4 & 72.9 & 84.4 & 82.0\\
 & CLoRA-minor-k1024 & 73.1 & 83.7 & 79.2 & 93.7 & 84.8 & 87.1 & 73.2 & 83.2 & 82.3\\
 & CLoRA-minor-k2048 & 72.9 & 84.2 & 80.8 & 93.7 & 85.3 & 87.2 & 73.5 & \textbf{86.0} & 83.0\\
\hline
LLaMA-3-8B & LoRA & 70.8 & 85.2 & 79.9 & 91.7 & 84.3 & 84.2 & 71.2 & 79.0 & 80.8\\
 & PiSSA & 67.1 & 81.1 & 77.2 & 83.6 & 78.9 & 77.7 & 63.2 & 74.6 & 75.4\\
 & MiLoRA & 68.8 & 86.7 & 77.2 & 92.9 & 85.6 & 86.8 & 75.5 & 81.8 & 81.9\\
 & DoRA & 74.6 & 89.3 & 79.9 & 95.5 & 85.6 & 90.5 & 80.4 & 85.8 & 85.2\\
\hline
 & CLoRA-random-k128 & 75.5 & 89.1 & 81.6 & 95.9 & 87.9 & 92.6 & 81.8 & 86.8 & 86.4\\
 & CLoRA-random-k256 & 75.3 & 88.8 & 81.4 & 85.7 & 88.7 & 92.7 & 82.3 & 88.4 & 85.4\\
 & CLoRA-random-k512 & 75.9 & 89.3 & 82.6 & 96.3 & \textbf{88.9} & 92.1 & 82.9 & 86.8 & 86.9\\
 & CLoRA-random-k1024 & \textbf{76.5} & 89.1 & 82.1 & 96.3 & 88.6 & 93.0 & 81.7 & \textbf{90.0} & 87.2\\
      & CLoRA-random-k2048 & 76.2 & \textbf{90.0} & \textbf{82.7} & \textbf{96.6} & 88.8 & \textbf{93.3} & \textbf{83.4} & 89.2 & \textbf{87.5}\\
    \hline
  \end{tabular}
  }
\end{center}
\caption{\label{table:full-in-domain}
  In-domain results on commonsense reasoning evaluations, with accuracy scores (\%) reported. \textbf{Bold} font indicates the highest performance for each dataset across the different PEFT methods for each base model.
  }
\end{table*}

\begin{table*}
  \begin{center}
    \begin{tabular}{llccc}
      \hline
      Model & PEFT & BBH & MMLU-Pro & Avg.\\
      \hline
      LLaMA-2-7B & - & 34.91 & 18.56 & 26.74\\
            & LoRA & 26.69 & 14.46 & 20.58\\
            & PiSSA & 29.54 & 11.33 & 20.44\\
            & MiLoRA & 25.14 & 17.74 & 21.44\\
            & DoRA & 28.24 & 11.67 & 19.96\\
      \hline
            & LoRA-r8 & 26.90 & 14.58 & 20.74\\
            & LoRA-r16 & 26.73 & 11.54 & 19.13\\
            & LoRA-L2-0.00001 & 32.93 & 16.59 & 24.76\\
      \hline
            & CLoRA-random-k128 & 30.82 & 12.07 & 21.45\\
            & CLoRA-random-k256 & 31.92 & 17.81 & 24.87\\
            & CLoRA-random-k512 & 34.32 & 17.00 & 25.66\\
            & CLoRA-random-k1024 & 36.49 & 19.52 & 28.01\\
            & CLoRA-random-k2048 & 38.67 & \textbf{20.59} & 29.63\\
      \hline
            & CLoRA-major-k128 & 32.69 & 18.09 & 25.39\\
            & CLoRA-major-k256 & 35.11 & 18.89 & 27.00\\
            & CLoRA-major-k512 & 35.81 & 19.88 & 27.85\\
            & CLoRA-major-k1024 & 37.06 & 19.73 & 28.40\\
            & CLoRA-major-k2048 & 38.83 & 20.08 & 29.46\\
      \hline
            & CLoRA-minor-k128 & 34.06 & 17.03 & 25.55\\
            & CLoRA-minor-k256 & 33.16 & 17.11 & 25.13\\
            & CLoRA-minor-k512 & 35.42 & 18.97 & 27.20\\
            & CLoRA-minor-k1024 & 37.08 & 18.87 & 27.98\\
            & CLoRA-minor-k2048 & \textbf{40.96} & 20.37 & \textbf{30.67}\\
      \hline
\end{tabular}
\end{center}
\caption{\label{table:full-out-domain}
  Out-domain results on two LLM benchmarks, with accuracy scores (\%) reported. \textbf{Bold} font indicates the highest performance for each benchmark across all methods.
  }
\end{table*}

We report the full results that we conducted on in-domain evaluation(Table~\ref{table:full-in-domain}) and out-domain evaluation(Table~\ref{table:full-out-domain}) for commonsense reasoning finetuning.
Results for LLaMA-3-8b are also included for CLoRA-random.
All models use LoRA rank $r$ of 32 unless specified otherwise.

\subsubsection{Analysis for different CLoRA variants}
Results indicate that the choice of regularization matrix does influence the effectiveness of CLoRA,
albeit not significantly.
Generally, we recommend using random initialization (CLoRA-random)
or initialization from major SVD components (CLoRA-major).

\section{Detailed Experiment Setups for Continual Learning}
\label{appendix:cont}
\begin{table*}
  \center
  \begin{tabular}{llll}
    \hline
    \textbf{Dataset name} & \textbf{Category} & \textbf{Task} & \textbf{Domain} \\ \hline
    Yelp                 & CL Benchmark      & sentiment analysis   & Yelp reviews       \\
    Amazon            & CL Benchmark      & sentiment analysis   & Amazon reviews     \\
    DBpedia          & CL Benchmark      & topic classification & Wikipedia          \\
    Yahoo            & CL Benchmark      & topic classification & Yahoo Q\&A         \\
    AG News          & CL Benchmark      & topic classification & news               \\
    \hline
    MNLI             & GLUE              & NLI                 & various            \\
    QQP              & GLUE              & paragraph detection  & Quora              \\
    RTE              & GLUE              & NLI                 & news, Wikipedia    \\
    SST-2            & GLUE              & sentiment analysis   & movie reviews      \\
    \hline
    WiC             & SuperGLUE         & word sense disambiguation & lexical databases  \\
    CB              & SuperGLUE         & NLI                 & various            \\
    COPA            & SuperGLUE         & QA                  & blogs, encyclopedia\\
    BoolQA          & SuperGLUE         & boolean QA          & Wikipedia          \\
    MultiRC         & SuperGLUE         & QA                  & various            \\
    IMDB            & SuperGLUE         & sentiment analysis  & movie reviews      \\
    \hline
    \end{tabular}
    \caption{\label{table:cont_data}Summary of datasets used in the continual learning setting.}
  \end{table*}

  \begin{table*}
    \centering
    \begin{tabular}{lll}
      \hline
      Order & Task Sequence \\
      \hline
      1 & dbpedia $\rightarrow$ amazon $\rightarrow$ yahoo $\rightarrow$ ag \\
      2 & dbpedia $\rightarrow$ amazon $\rightarrow$ ag $\rightarrow$ yahoo \\
      3 & yahoo $\rightarrow$ amazon $\rightarrow$ ag $\rightarrow$ dbpedia \\
      \hline
      4  & mnli $\rightarrow$ cb $\rightarrow$ wic $\rightarrow$ copa $\rightarrow$ qqp $\rightarrow$ boolqa $\rightarrow$ rte $\rightarrow$ imdb \\
            & $\rightarrow$ yelp $\rightarrow$ amazon $\rightarrow$ sst-2 $\rightarrow$ dbpedia $\rightarrow$ ag $\rightarrow$ multirc $\rightarrow$ yahoo \\
      5 & multirc $\rightarrow$ boolqa $\rightarrow$ wic $\rightarrow$ mnli $\rightarrow$ cb $\rightarrow$ copa $\rightarrow$ qqp $\rightarrow$ rte\\
      & $\rightarrow$ imdb $\rightarrow$ sst-2 $\rightarrow$ dbpedia $\rightarrow$ ag $\rightarrow$ yelp $\rightarrow$ amazon $\rightarrow$ yahoo \\
      6 & yelp $\rightarrow$ amazon $\rightarrow$ mnli $\rightarrow$ cb $\rightarrow$ copa $\rightarrow$ qqp $\rightarrow$ rte $\rightarrow$ imdb \\
      &$\rightarrow$ sst-2 $\rightarrow$ dbpedia $\rightarrow$ ag $\rightarrow$ yahoo $\rightarrow$ multirc $\rightarrow$ boolqa $\rightarrow$ wic \\
      \hline
    \end{tabular}
    \caption{\label{table:cont_order}Order of finetuning in the continual learning setting.}
  \end{table*}

\subsection{Dataset Details}
We list the details of the datasets used in Table~\ref{table:cont_data}.
Order of finetuning are listed in Table~\ref{table:cont_order}.
\subsection{Computation Environment}
All of our experiments are conducted on 1 NVIDIA GeForce RTX 3090 GPU.
All methods for LoRA subsequents use Huggingface peft library,
training is conducted using trainer in Huggingface transformers library,
with DeepSpeed ZeRO intergration.
\subsection{Comparision Methods}

Here we provide details for continual learning baselines for our continual learning experiment setting.
\begin{itemize}[leftmargin=*]
  \item \textbf{Replay:} data-based method that replay samples from old tasks when learning new tasks to avoid forgetting.
  \item \textbf{L2P:} architecture-based method that uses the input to dynamically select and update prompts from the prompt pool in an instance-wise fashion.
  \item \textbf{LFPT5:} architecture-based method that continuously train a soft prompt that simultaneously learns to solve the tasks and generate training samples for replay.
  \item \textbf{EWC:} learning-based method that finetune the whole model with a regularization loss that prevents updating parameters that could interfere with previously learned tasks.
  \item \textbf{LwF:} learning-based method that constrains the shared representation layer to be similar to its original state before learning the new task.
  \item \textbf{O-LoRA:} architecture and learning-based method that prevent subsequent LoRA update interfere previous.
\end{itemize}
\end{document}